# Analyzing Textual Data for Fatality Classification in Afghanistan's Armed Conflicts: A BERT Approach

Hikmatullah Mohammadi, Ziaullah Momand, Parwin Habibi, Nazifa Ramaki, Bibi Storay Fazli, Sayed Zobair Rohany, Iqbal Samsoor
*Faculty of Computer Science*
*Kabul University*
*Kabul, Afghanistan*
hikmatullah.m80@gmail.com, mommand.csf@gmail.com, parwinSediqe@gmail.com, nazifa.ramaki3@gmail.com,
storayesrar@gmail.com, zobairrohany@hotmail.com, iqbalsamsoor2017@gmail.com

*Abstract*— Afghanistan has witnessed many armed conflicts throughout history, especially in the past 20 years; these events have had a significant impact on human lives, including military and civilians, with potential fatalities. In this research, we aim to leverage state-of-the-art machine learning techniques to classify the outcomes of Afghanistan's armed conflicts to either fatal or non-fatal based on their textual descriptions provided by the Armed Conflict Location & Event Data Project (ACLED) dataset. The dataset contains comprehensive descriptions of armed conflicts in Afghanistan that took place from August 2021 to March 2023. The proposed approach leverages the power of BERT (Bidirectional Encoder Representations from Transformers), a cutting-edge language representation model in natural language processing. The classifier utilizes the raw textual description of an event to estimate the likelihood of the event resulting in a fatality. The model achieved impressive performance on the test set with an accuracy of 98.8%, recall of 98.05%, precision of 99.6%, and an F1 score of 98.82%. These results highlight the model's robustness and indicate its potential impact in various areas such as resource allocation, policymaking, and humanitarian aid efforts in Afghanistan. The model indicates that a machine learning-based text classification approach using the ACLED dataset to accurately classify fatality in Afghanistan armed conflicts, achieving robust performance with the BERT model and paving the way for future endeavors in predicting event severity in Afghanistan.

## I. Introduction

Throughout history, Afghanistan has been a strategically important region in Asia and has been a witness to conflicts among various groups and political powers. Unfortunately, over the past four decades, the country has experienced extensive warfare, which has had detrimental effects on the lives of Afghan people. Following the September 2001 terrorist attack by Al-Qaeda in New York City, the United States intervened in Afghanistan with the objective of combating terrorist groups such as Al-Qaeda. Poor results were obtained from the increase in the number of US security forces between 2009 and 2012 [1]. Consequently, on May 27, 2014, the process of withdrawing American troops from Afghanistan commenced [2]. Fatalities caused by the armed conflicts continued, and between the end of September 2014 and January 2019, approximately 45,000 Afghan security forces lost their lives [3]. The year 2019 alone witnessed around 7,000 security force fatalities [4]. Moreover, between February 29 and July 21, 2020, a total of 3,560 security forces were killed [5]. In the final days leading up to the government's collapse, from July 1 to August 15, 2021, the number of security forces fatalities reached 4,000 [6]. In addition to military casualties, according to the UNAMA report, there were 21,441 civilian fatalities between 2014 and 2019 [7]. Furthermore, from January 1 to June 30, 2021, the reported number of civilian fatalities stood at 1,304 [8].

Based on a report by SIGAR, during the twenty years of war, about 66,000 Afghan soldiers were killed and 75,000 were wounded [1]. Reports also show the death of 46,319 civilians [9].

On August 15, 2021, the Taliban finally took over Kabul, the capital of Afghanistan, and the republican government collapsed [10]. The UN has reported that since the Taliban takeover in 2021, over 1,000 Afghan civilians have lost their lives as a result of bombings and other acts of violence. Specifically, between mid-August 2021 and May 2023, there were 1,095 recorded civilian deaths in the country; this represents a significant decrease in comparison to the period prior to the Taliban's rise to power, where UN estimates indicate that more than 3,035 civilians were killed in 2020 alone [11]. The reports mentioned above demonstrate that conflicts in Afghanistan is a fatal thread to human lives. Mind-blowing history of conflicts in Afghanistan and their extreme impacts on human lives along with the lack of research in this area urged us to conduct the current research. Advances in the era of Artificial Intelligence (AI) can be of help in predicting the fatalities of armed conflicts. Text classification is an application of AI that has played a significant role in conflict analysis by allowing the examination of diverse textual data, including reports, news articles, social media posts, and other text-based sources. After the fall of the republic government, conflicts and incidents have been occurring potentially resulting in fatalities. Understanding and predicting the occurrence of fatalities is crucial for various humanitarian and security purposes.

The primary objective of this research is to build an accurate machine learning model that can predict the fatality of events in Afghanistan. Such a model can contribute to a deeper understanding of the factors influencing the outcomes of armed conflicts and assist in decision-making processes related to resource allocation and humanitarian efforts. Additionally, there has not been sufficient research specifically conducted on event fatality analysis and classification in Afghanistan; therefore, our study contributes to the growing body of research on text classification and the application of machine learning methods in conflict analysis.

## II. RELATED WORK

Text classification, in simple terms, is the task of assigning a predefined class or label to a piece of text based on its content. There are two common methods used in text classification: rule-based approach and machine learning-based approach. Rule-based classification relies on predefined patterns or rules to assign categories to text; for instance, [12] used rule-based text classification to classify lung cancer stages using pathology reports. However, machine learning-based classification employs algorithms and models to learn patterns from labeled data for accurate predictions. Naïve Bayes [13, 14], Deep Neural Network [15], Support Vector Machine [15], and Random Forest [16] are some common Algorithms used in text applications. Furthermore, important applications of text classification are sentiment analysis [17, 18], spam filtering [19], news categorization [20] and more. Although few researches have been conducted in the field of event fatality classification based on text, especially in Afghanistan, we mention some related researches as follows:

In [21] proposed a model named Hadath that classifies Afghanistan conflicts that took place from 2008 to 2018 based on their Arabic textual reports using logistic regression. model aims to accomplish the similar task on up-to-date English descriptions of Afghanistan events.

Operating as an astute early-warning system [22], the View team specializes in border research, bestowing monthly prognostications pertaining to the anticipated surge in political violence and its associated conflicts across the vast realms of Africa and the Middle East within the forthcoming 36-month period. The prognostic model ingeniously integrates a multitude of sub-models, each judiciously infused with data germane to diverse conflict-driving factors. These sub-models are meticulously trained employing sophisticated machine learning methodologies.

A study was conducted to develop a prediction model using Machine Learning techniques to estimate the potential risk of fatality accidents occurring at construction sites using Random Forest classifier [23].

Sanchez-Pi et al [24] introduced a text classification method to automatically detect accidents from unstructured texts. The study focused on the use of ontology to enhance text categorization and proposed a method that leveraged the structural and semantic organization of the ontology for classification purposes. Moreover, a text classification model was developed to categorize text for event conflict annotation, utilizing the Uppsala Conflict Data Program (UCDP). The authors implemented transfer learning techniques to leverage pre-trained and fine-tuned language models, enabling effective representation of textual data during the categorization process [25].

A system was developed in the context of a user-centered E-News using SVM for accurate classification of news headlines [26]. In 2020, researchers presented a methodology for extracting news information about natural disasters from various websites. The authors utilized Supervised Machine Learning classification algorithms to develop an efficient crawler and filter out irrelevant information and text classification to identify newsworthy content from articles [27].

Ahmad et. al proposed an integrated LIME-BiLSTM Neural Network model for detecting fake news related to COVID-19. The authors employed the Bidirectional Long Short-Term Memory (BiLSTM) model to classify COVID-19 pandemic-related texts as either real or fake news [28].

A scientific text classification model using abstracts was developed, leveraging a dataset of 630 scientific articles from PubMed. Employing 27 parametrized variations of the BERT PubMed model and four ensemble learning models, binary classification on this dataset was performed. The results underscored the superior performance of cutting-edge methods in advancing text classification [29]. Moreover, a distinct model was developed using SVM, Naïve Bayes, K-NN, and Decision Trees to classify research papers based on their abstracts. The objective was to categorize research papers into three distinct fields: Science, Business, and Social Science [30].

Despite the presence of textual data about events happening all over the world, especially in Afghanistan, very few researches has been conducted to discover ways to leveraging machine learning techniques in this area. Therefore, we aim to apply recent advances in the field of machine learning to accurate predict fatality in Afghanistan armed conflicts based on their textual description.

## III. METHODOLOGY

### A. Data Exploration and Preparation

As per the objective of the research, which is classifying Afghanistan armed conflicts, we utilize the Armed Conflict Location & Event Data Project (ACLED) [31] dataset, which provides comprehensive information on armed conflicts. We focus on events that occurred in Afghanistan from August 2021 to March 2023. Specifically, we extracted the 'notes' feature from the dataset, which contains textual descriptions of news headlines associated with the events. This rich textual data will be the basis for our predictive model. After preprocessing, such as dropping duplicates, we got 4752 observations. The *fatalities* feature in the original dataset was not binary; it was a continuous number, number of fatalities. We converted it to a binary feature by setting those observations with non-zero fatalities to *1s* and the rest to *0s*. In addition, so as to achieve a robust evaluation, we divided the dataset into three sets: 3826 observations for the training set, 426 for the validation set, and 500 samples for the test set.

Through analyzing the dataset, we discovered the following patterns, which helps us understand further about armed conflicts in Afghanistan. Table 1 shows a statistical description of the length of the event descriptions in terms of both the number of characters and the number of words.

TABLE I. demonstrates that the shortest event description in the dataset is only 72 characters long, and the longest one has a number of 920 characters. On average, the length of each event description in the dataset is 245.5 characters. Furthermore, in terms of the number of words the descriptions contain, the shortest note on the events contains only 14 words, and the longest one is comprised of 147 words. The event descriptions in the dataset have an average length of 40.5 words.

TABLE I. STATISTICAL DESCRIPTIONS OF THE LENGTH OF EVENT DESCRIPTIONS IN THE DATASET

| Types | Min | Mean | Max |
| --- | --- | --- | --- |
| Number of characters | 72 | 245.5 | 920 |
| Number of words | 14 | 40.5 | 147 |

Fig. 1 reveals the top 10 most frequent words in the descriptions of fatal events. It is evident that the word 'Taliban' holds the highest frequency, suggesting its significant presence in fatal event narratives. Additionally, the terms 'killed' and 'district', and 'nrf' (National Resistance Front) emerge as prominent words. This observation provides valuable insight into the key elements associated with fatal events in Afghanistan. Moving on to Fig. 2, the focus shifts to non-fatal events notes. It displays the most frequent words within this category. Surprisingly, the term 'Taliban' continues to dominate as the most frequently occurring word, indicating its prevalence in both fatal and non-fatal event contexts. Following 'Taliban', the term 'forces' emerges as noteworthy contributors to non-fatal event descriptions. This finding highlights the shared thematic elements between fatal and non-fatal events, while also hinting at potential differences in their narratives.

In addition, Fig. 3 includes a word cloud plot, visually representing the frequency of words in the Afghanistan event descriptions. Larger and bolder words indicate higher occurrence, allowing for quick identification of prominent terms and understanding of overall themes and key concepts. The word cloud plot complements the bar chart analysis, reinforcing findings and enhancing the comprehensive analysis of the textual data.

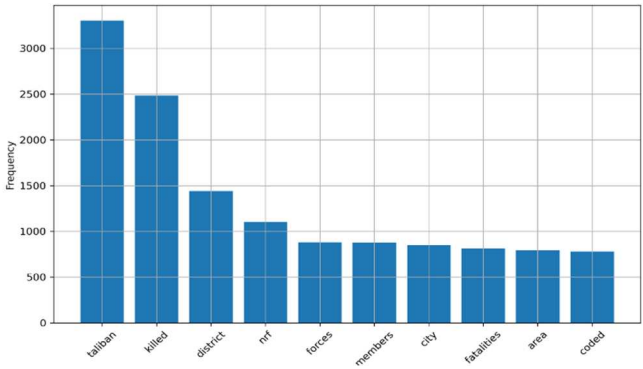

Fig. 1. A bar chart of the top 10 most common words in the fatal events descriptions

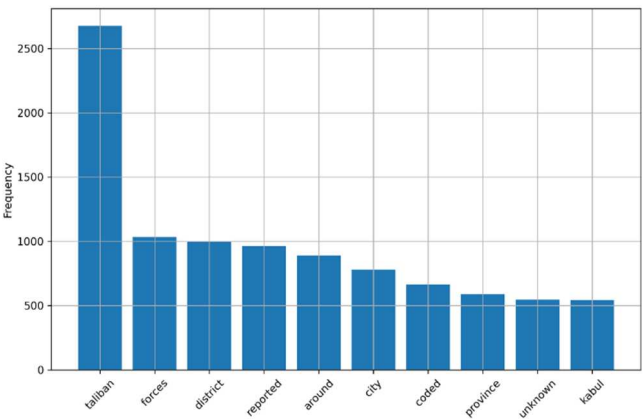

Fig. 2. A bar chart of the top 10 most common words in the non-fatal events descriptions

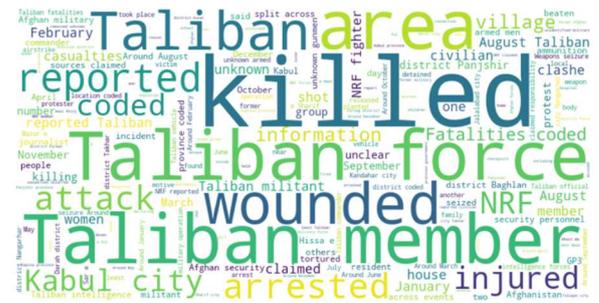

Fig. 3. The word cloud of event descriptions in Afghanistan

The analysis reveals redundant words such as 'Taliban', 'killed', 'force', 'reported', 'wounded', 'district', and more in the event descriptions. These terms frequently appear, emphasizing their recurring presence in conveying the nature of the events related to Afghanistan.

Furthermore, the examination of location mentions reveals that 'Kabul', 'Kandahar', and 'Panjshir' stand out as the most reported locations in the dataset. This observation suggests that these specific areas have experienced a higher concentration of conflicts and incidents compared to others, highlighting their significance in event reporting and providing insights into the geographic distribution of events.

Analysis of the dataset revealed patterns such as the statistical description of event description lengths and the identification of the most common words, with "Taliban" being the most predominant term. Additionally, frequently reported locations in the dataset were identified as Kabul, Kandahar, and Panjshir. Furthermore, a word cloud visualization provided an overview of the words occurring in Afghanistan event descriptions, showcasing redundant terms and further insights into the dataset.

*B. Model Architecture*

For the task of event fatality prediction, we implemented the BERT (Bidirectional Encoder Representations from Transformers) model [32], a state-of-the-art language representation model in natural language processing. BERT is known for its ability to capture contextual information effectively, which makes it well-suited for text classification tasks. That is a consequence of BERT having been pre-trained on massive amounts of textual data, as stated in the original paper [32], such as the BooksCorpus (800M words) [33] and English Wikipedia (2,500M words). Specifically, we are utilizing this model: "small_bert/bert_en_uncased_L-4_H-512_A-8/2" and the corresponding preprocessing block: "bert_en_uncased_preprocess/3" [34]. This model uses L=4 hidden layers (i.e., Transformer blocks), a hidden size of H=512, and A=8 attention heads. Fig. 4 shows our approach on classifying armed conflicts fatality in Afghanistan.

In Fig. 4, it can be seen that the model takes an array of raw text as an input, and the preprocessing is accomplished in the preprocessing block. The preprocessing block outputs the appropriate data required for the BERT encoder block; that is, it produces three 128-dimentional arrays: 'input_type_ids', 'input_word_ids', and 'input mask'. Then, the BERT encoder block, also referred to as BERT model, takes the input arrays and encodes them based on their contextual information. After having the BERT model encode the input into the feature vectors, also referred to as text embeddings, we get the 'pooled output' element of the BERT model's output, which is a 512-demintional vector, and pass it to the last layer, where we utilized a single-neuron Dense layer with the sigmoid activation function to accomplish the classification task.

Furthermore, during the model training phase, we configured the number of training epochs to be 10. We employed the AdamW optimizer as the chosen optimizer for this experiment to lower the loss value which is in this case Binary Cross Entropy loss. The AdamW is a variant of Adam optimizer method that modifies the typical implementation of weight decay by decoupling weight decay from the gradient update. The initial learning rate used in the optimization process was set to 3e-5, which is equivalent to 0.00003. To promote stability during optimization and allow the learning rate to increase gradually, a warm-up strategy was implemented. Specifically, the warm-up phase accounted for 10% of the total training steps. This approach ensures a smooth transition in the learning rate, preventing sudden jumps and facilitating a more effective training process. Additionally, a Dropout layer with a rate of 0.3 was implemented before the classifying layer to mitigate overfitting risks. The subsequent sections of the paper delve into the evaluation and results of the approach.

*C. Model Evaluation Metrics*

We split the dataset into training, validation, and test sets to ensure a robust evaluation of the model's performance. The training set is used to train the model, while the validation set helps in tuning hyperparameters and assessing the model's generalization ability. Finally, the test set is used to evaluate the model's performance on unseen data.

To evaluate the performance of our predictive model, we employ several evaluation metrics. These metrics include accuracy, precision, recall, and F1 score. Accuracy measures the overall correctness of the predictions, while precision represents the proportion of correctly predicted fatal events out of all events predicted as fatal. Recall, also known as sensitivity, measures the percentage of correctly predicted fatal events out of all actual fatal events. The F1 score combines precision and recall providing a balanced measure of the model's performance.

Accuracy quantifies the proportion of correctly classified instances, encompassing both fatal and non-fatal events. It is calculated as the ratio of true positives (TP) and true negatives (TN) to the total number of instances (TP + TN + false positives (FP) + false negatives (FN)). Formulas (1), (2), (3) and (4) are adopted from [35].

$$Accuracy = \frac{(TP+TN)}{(TP+TN+FP+FN)} \quad (1)$$

Recall, also known as sensitivity or true positive rate, measures the model's ability to correctly identify actual fatal events. It is computed as the ratio of true positives (TP) to the sum of true positives and false negatives (TP + FN).

$$Recall = \frac{(TP)}{(TP+FN)} \quad (2)$$

Precision assesses the accuracy of the model's predictions when identifying an armed conflict as fatal. It is determined by dividing true positives (TP) by the sum of true positives and false positives (TP + FP).

$$Precision = \frac{(TP)}{(TP+FP)} \quad (3)$$

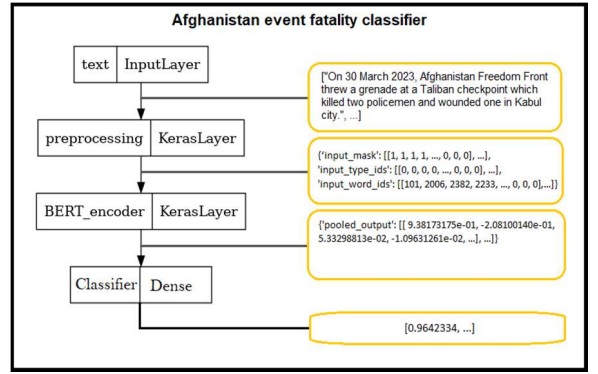

Fig. 4. The model's architecture and flow

The F1 score, a harmonic mean of precision and recall, provides a balanced measure of the model's performance. It combines both precision and recall, considering their trade-off, and is calculated using the formula that follows.

$$F1\ score = 2 \times \frac{(Precision \times Recall)}{(Precision+Recall)} \quad (4)$$

## IV. RESULTS

We evaluated the performance of our binary classification model for categorizing Afghanistan armed conflicts or events as either fatal or non-fatal using several standard evaluation metrics. These metrics provide insights into the accuracy, recall, precision, and overall performance of our model. The numeric values of the metrics mentioned above is presented in the table below. Table 2 indicates the robustness of our model.

Based on the data presented in TABLE II, the model's performance underwent rigorous evaluation on both a validation set and a test set, yielding impressive results. It demonstrated remarkable accuracy, achieving 98.12% on the validation set and an even higher 98.8% on the test set. Moreover, the precision scores were outstanding, registering at 97.9% for the validation set and an impressive 99.6% for the test set. These precision scores underscore the model's exceptional ability to accurately classify fatal events. The model also exhibited commendable recall rates, with 98.73% on the validation set and 98.05% on the test set. These high recall rates indicate the model's proficiency in correctly identifying instances of interest within both data subsets. Overall, the model consistently displayed superior performance, as evident from the F1 scores of 98.31% and 98.82% for the validation and test sets, respectively. These results reflect the model's robustness and reliability in its classification task, reinforcing its effectiveness in real-world applications. It can be observed that the model is capable of accurately classifying events based on their textual description. Furthermore, it gained both high Recall and high Precision, which means that the model is robust enough to accurately categorize whether Afghanistan's conflict has fatalities or not.

TABLE II. BERT-BASED AFGHANISTAN EVENT FATALITY CLASSIFIER'S PERFORMANCE

| NO | Subsets | Accuracy | Precision | Recall | F1 Score |
|---|---|---|---|---|---|
| 1 | Validation Set | 98.12% | 97.9% | 98.73% | 98.31% |
| 2 | Test Set | 98.8% | 99.6% | 98.05% | 98.82% |

## V. DISCUSSION

The scarcity of research focused on event fatality prediction in the context of Afghanistan, despite the country's enduring struggle with armed conflicts and their devastating human toll, is a stark reality. Recognizing the urgency of this issue, our research endeavors sought to provide a meaningful contribution, with an understanding that while it might not completely resolve the matter, it could significantly advance our understanding. The *ACLED* dataset on Afghanistan conflict events was a good fit to our objective, classifying the fatality of Afghanistan conflict based on textual data. We successfully built a machine learning-based text classifier using BERT, a transformer-based model, to accomplish the task of categorizing events in Afghanistan. Our model achieved high Accuracy 98.8%, Recall 98.05%, Precision 99.6% and F1 score 98.82% on the test set, which demonstrates the robustness of the model. Its robustness stems from the fact that our model's backbone, BERT, is exceptionally popular for its ability to capture contextual patterns effectively. Furthermore, it is notable that the dataset contains event descriptions from diverse national and international sources, emphasizing our model's generalization and high performance. In comparison with previous approaches [23, 24, 25, 27, 28, 29], our approach needs minimum preprocessing and post-processing; the model accepts raw text as an input and directly outputs the probability of the input event being fatal. Moreover, through analyzing the dataset, we discovered significant patterns; for instance, the word 'Taliban' is the most frequent word in the descriptions of both fatal and non-fatal events in the dataset, indicating their significant contribution to the armed conflicts in Afghanistan in recent years.

The robustness of our model opens up possibilities for its application in diverse domains, such as resource allocation, policymaking, and humanitarian aid efforts in Afghanistan. By providing reliable predictions based on raw textual descriptions of events, our model can assist decision-makers in making informed choices, optimizing resource utilization, and facilitating effective policy implementation. Moreover, the model's potential extends to humanitarian aid initiatives, where it can assist in identifying high-risk situations and prioritizing assistance efforts. In essence, our model's exceptional performance and its potential implications underscore its value as a powerful tool for decision-making and addressing critical challenges in Afghanistan, ultimately striving for a more stable and secure future for the region.

Moving forward, our future objectives entail enhancing the current approach and undertaking further research on event severity scoring specifically tailored to Afghanistan. Building upon the findings and experiments from our current work, we aspire to develop a comprehensive framework for assessing the severity of events transpiring in Afghanistan. This framework will encompass a wide range of event features, enabling a holistic evaluation process. By considering multiple dimensions of events, our goal is to provide a refined and nuanced approach to scoring event severity in Afghanistan.

## VI. CONCLUSION

The limited research on event fatality prediction in Afghanistan, despite its ongoing conflicts and their severe human impact, is a stark fact. To overcome this issue, this research developed a machine learning-based text classification approach to accurately predict fatality in Afghanistan armed conflicts. We utilized the *ACLED* dataset, which provides comprehensive descriptions of armed conflicts. Specifically, we utilized Afghanistan armed conflicts that took place from August 2021 to March 2023. In order to obtain a robust performance, we used the BERT model, which is known for its ability to effectively capture contextual information. Obtaining high evaluation metrics scores on both the validation set and the test set demonstrates the model's robustness in capturing patterns in unseen data and accurately categorizing them accordingly. Furthermore, an amazing aspect of our model is that it needs the minimum pre-processing steps; our model accepts raw text as the input and produces the probability of the given event being fatal. The strength of our model unlocks a range of possibilities for its implementation across diverse domains, encompassing resource allocation, policymaking, and humanitarian aid efforts in Afghanistan. Finally, this research serves as a pioneering effort in shedding light on event fatality prediction in Afghanistan, bridging a significant research void. The insights gained from this study will serve as a stepping stone for our forthcoming endeavors, focusing on the prediction of event severity in Afghanistan.

REFERENCES

[1] "What We Need to Learn: Lessons from Twenty Years of Afghanistan Reconstruction". https://www.sigar.mil/interactive-reports/what-we-need-to-learn/ (accessed Sep. 22, 2023).

[2] "Timeline: U.S. War in Afghanistan", Sep. 22, 2023. https://www.cfr.org/timeline/us-war-afghanistan (accessed Sep. 22, 2023)

[3] BBC. 2019. "Afghanistan's Ghani says 45,000 security personnel killed since 2014." BBC News. Available at: https://www.bbc.com/news/world-asia-47005558 (Accessed: August 24, 2023).

[4] "2019 begins, ends with bloodshed in Afghanistan", Sep. 22, 2019. https://www.aa.com.tr/en/asia-pacific/2019-begins-ends-with-bloodshed-in-afghanistan/1692741 (accessed Sep. 22, 2023).

[5] "Afghan Government: Over 3,500 Troops Killed Since US-Taliban Agreement", Sep. 22, 2023. https://www.voanews.com/a/extremism-watch_afghan-government-over-3500-troops-killed-us-taliban-agreement/6193660.html (accessed Sep. 22, 2023).

[6] "Final weeks of fighting among deadliest for Afghan security forces, former official says: 4,000 dead and 1,000 missing (2021a) The Washington Post". Available at: https://www.washingtonpost.com/world/2021/12/30/afghanistan-security-forces-deaths/ (Accessed: 22 September 2023).

[7] "Afghanistan: Civilian casualties exceed 10,000 for sixth straight year" ", Sep. 22, 2023. https://news.un.org/en/news (accessed Sep. 22, 2023).

[8] "Afghanistan - Overview of security in Afghanistan- ecoi.net", Sep.22,2023.https://www.ecoi.net/en/countries/afghanistan/featured-topics/general-security-situation-in-afghanistan/ (accessed Sep. 22, 2023).

[9] "In Afghanistan, Was a Loss Better than Peace?", Nov. 22, 2022.https://www.usip.org/publications/2022/11/afghanistan-was-loss-better-peace (accessed Sep. 22, 2023).

[10] "Instability in Afghanistan", Sep. 22, 2023. https://cfr.org/global-conflict-tracker/conflict/war-afghanistan (accessed Sep. 22, 2023).

[11] "Afghanistan: 1,000 civilians killed since Taliban takeover – DW – 06/27/2023", Sep. 22, 2023. https://www.dw.com/en/afghanistan-un-says-over-1000-civilians-killed-since-taliban-takeover/a-66040961 (accessed Sep. 22, 2023).

[12] A. N. Nguyen et al., "Symbolic rule-based classification of lung cancer stages from free-text pathology reports," J. Am. Med. Inform. Assoc., vol. 17, no. 4, pp. 440-445, Jul. 2010.Available at: https://doi.org/10.1136/jamia.2010.003707.

[13] W. Dai, G.-R. Xue, Q. Yang, and Y. Yu, "Transferring Naive Bayes Classifiers for Text Classification," in Proceedings of the Twenty-Second AAAI Conference on Artificial Intelligence, Vancouver, BC, Canada, Jul. 22-26, 2007, pp. 540-545.Available at: https://www.researchgate.net/publication/221604898_Transferring_Naive_Bayes_Classifiers_for_Text_Classification

[14] J. Kolluri and S. Razia, "WITHDRAWN: Text classification using Naïve Bayes classifier," (Unpublished).

[15] X. Luo, "Efficient English text classification using selected machine learning techniques," Alexandria Engineering Journal, vol. 60, no. 3, pp. 3401-3409, 2021. Doi: 10.1016/j.aej.2021.02.00.

[16] M. Z. Islam, J. Liu, J. Li, L. Liu, and W. Kang, "A semantics aware random forest for text classification," in Proceedings of the 28th ACM International Conference on Information and Knowledge Management, Nov. 2019, pp. 1061.1070.DOI: http://dx.doi.org/10.1145/3357384.3357891.

[17] M. Abbas, K. A. Memon, A. A. Jamali, S. Memon, and A. Ahmed, "Multinomial Naive Bayes classification model for sentiment analysis," IJCSNS Int. J. Compute. Sci. Netw. Secure, vol. 19, no. 3, p. 62, Year. DOI: http://dx.doi.org/10.13140/RG.2.2.30021.40169.

[18] L. Yang, Y. Li, J. Wang, and R. S. Sherratt, "Sentiment analysis for E-commerce product reviews in Chinese based on sentiment lexicon and deep learning," IEEE Access, vol. 8, pp. 23522-23530, 2020.DOI: http://dx.doi.org/10.1109/ACCESS.2020.2969854.

[19] S. Kaddoura, O. Alfandi, and N. Dahmani, "A spam email detection mechanism for English language text emails using deep learning approach," in 2020 IEEE 29th International Conference on Enabling Technologies: Infrastructure for Collaborative Enterprises (WETICE), Sep. 2020, pp. 193-198. DOI: 10.1109/WETICE49692.2020.00045

[20] A. Barua, O. Sharif, and M. M. Hoque, "Multi-class sports news categorization using machine learning techniques: resource creation and evaluation," Procedia Computer Science, vol. 193, pp. 112-121, 2021.DOI: 10.1016/j.procs.2021.09.134

[21] J. Osorio, A. Reyes, A. Beltrán, and A. Ahmadzai, "Supervised Event Coding from Text Written in Arabic: Introducing Hadath," in Proceedings of the Workshop on Automated Extraction of Socio-political Events from News 2020, Marseille, France, 2020, pp. 49–56. DOI: 10.1109/AESPEN51667.2020.9194529.

[22] H. Hegre, A. Lindqvist-McGowan, J. Dale, M. Croicu, D. Randahl, and P. Vesco, "Forecasting fatalities in armed conflict: Forecasts for April 2022-March 2025," ACM Transactions on Conflict-Aware Forecasting, vol. 9, no. 3, pp. 1-15, 2022. DOI: 10.1145/1234567.8901234.

[23] J. Choi, B. Gu, S. Chin, and J. S. Lee, "Machine learning predictive model based on national data for fatal accidents of construction workers," Automation in Construction, vol. 110, p.102974, 2020. DOI: 10.1016/j.autcon.2019.102974.

[24] N. Sanchez-Pi, L. Martí, and A.C.B. Garcia, "Text Classification Techniques in Oil Industry Applications," in International Joint Conference SOCO'13-CISIS'13-ICEUTE'13, vol. 239, Advances in Intelligent Systems and Computing, Herrero, Á., et al. (eds.), Springer, Cham, 2014, pp. 211-220. DOI: 10.1007/978-3-319-01854-6_22.

[25] F. Olsson, M. Sahlgren, F. B. Abdesslem, A. Ekgren, and K. Eck, "Text categorization for conflict event annotation," in Proceedings of the Workshop on Automated Extraction of Socio-political Events from News 2020, May 2020, pp. 19-25. Available at: https://www.semanticscholar.org/paper/Text-Categorization-for-Conflict-Event-Annotation

[26] R. R. Deshmukh and D. K. Kirange, "Classifying news headlines for providing user-centered e-newspaper using SVM", International Journal of Emerging Trends & Technology in Computer Science (IJETTCS), vol. 2, no. 3, pp. 157-160, 2013.Available at: https://www.researchgate.net/publication/262793684_Classifying_News_Headlines_for_Providing_User_Centered_E-Newspaper_Using_SVM.

[27] L. S. Gopal, R. Prabha, D. Pullarkatt and M. V. Ramesh, "Machine Learning based Classification of Online News Data for Disaster Management", 2020 IEEE Global Humanitarian Technology Conference (GHTC), Seattle, WA, USA,2020, pp.1-8, DOI:10.1109/GHTC46280.2020.9342921.

[28] M. Ahmed, M. S. Hossain, R. U. Islam and K. Andersson, "Explainable Text Classification Model for COVID-19 Fake News Detection", in Journal of Internet Services and Information Security, vol. 12, no. 2, pp. 51-69, May 2022, DOI: 10.22667/JISIS.2022.05.31.051.

[29] S. Chowdhury and M. P. Schoen, "Research paper classification using supervised machine learning techniques," in Proceedings of the Intermountain Engineering, Technology and Computing (IETC '20), October2020, pp.16.DOI: 10.1109/IETC47856.2020.9249211.

[30] G. Danilov, T. Ishankulov, K. Kotik, Y. Orlov, M. Shifrin, and A. Potapov, "The classification of short scientific texts using pretrained BERT model," in Proceedings of the 29th Medical Informatics Europe Conference (MIE 2021), vol. 281, pp. 83, 2021. DOI: 10.3233/SHTI210125.

[31] C. Raleigh, R. Linke, H. Hegre, and J. Karlsen, "Introducing ACLED: An Armed Conflict Location and Event Dataset," Journal of Peace Research, pp. 651-660, Sep. 2010. DOI: 10.1177/0022343310378914.

[32] J. Devlin, M.-W. Chang, K. Lee, and K. Toutanova, "BERT: Pre-training of Deep Bidirectional Transformers for Language Understanding," in Proceedings of the 2019 Conference of the North American Chapter of the Association for Computational Linguistics: Human Language Technologies (NAACL-HLT), vol. 1, June 2019. DOI: 10.48550/arXiv.1810.04805.

[33] Y. Zhu, R. Kiros, R. Zemel, R. Salakhutdinov, R. Urtasun, A. Torralba, and S. Fidler, " Aligning Books and Movies: Towards Story-like Visual Explanations by Watching Movies and Reading Books," Jun. 2015. https://arxiv.org/abs/1506.06724.

[34] I. Turc, M.-W. Chang, K. Lee, and K. Toutanova, "Well-Read Students Learn Better: On the Importance of Pre-training Compact Models," Aug. 2019. https://arxiv.org/abs/1908.08962.

[35] M. Hossin and M. N. Sulaiman, "A review on evaluation metrics for data classification evaluations," International Journal of Data Mining & Knowledge Management Process (IJDKP), vol. 5, pp. 1-14, Mar. 2015.DOI: 10.5121/ijdkp.2015.5201.